\documentclass[conference]{IEEEtran}
\IEEEoverridecommandlockouts
\usepackage{cite}
\usepackage{amsmath}
\usepackage{amssymb}
\usepackage{amsfonts}
\usepackage{algorithmic}
\usepackage{graphicx}
\usepackage{textcomp}
\usepackage{stfloats}
\usepackage{placeins}
\usepackage{hyperref}
\usepackage{tabularx}
\usepackage{multicol}
\usepackage{array}
\usepackage{multirow}
\usepackage{pgfplotstable}
\usepackage{booktabs}
\usepackage{float}
\usepackage{subfig}
\usepackage{pgfplots}
\usepackage{comment}
\usepackage{footnote}
\pgfplotsset{compat=1.16}
\usepgfplotslibrary{groupplots}
\usetikzlibrary{patterns}
\definecolor{myblue}{RGB}{31,119,180}
\definecolor{myorange}{RGB}{255,127,14}
\definecolor{myteal}{RGB}{23,190,207}

\def\BibTeX{{\rm B\kern-.05em{\sc i\kern-.025em b}\kern-.08em
    T\kern-.1667em\lower.7ex\hbox{E}\kern-.125emX}}

\title{Explanations Leak: Membership Inference with Differential Privacy and Active Learning Defense}

\begin{document}
\author{\IEEEauthorblockN{Fatima Ezzeddine\textsuperscript{†, *}, Osama Zammar \textsuperscript{‡}, Silvia Giordano\textsuperscript{†} and Omran Ayoub\textsuperscript{†}}
\IEEEauthorblockA{
\textit{\textsuperscript{†} University of Applied Sciences and Arts of Southern Switzerland, Lugano, Switzerland}\\
\textit{\textsuperscript{*} Università della Svizzera italiana, Lugano, Switzerland}\\
\textit{\textsuperscript{‡} Lebanese University, Faculty of Science, Beirut, Lebanon}
}
}
\maketitle
\begin{abstract}
Counterfactual explanations (CFs) are increasingly integrated into Machine Learning as a Service (MLaaS) systems to improve transparency; however, ML models deployed via APIs are already vulnerable to privacy attacks such as membership inference and model extraction, and the impact of explanations on this threat landscape remains insufficiently understood. In this work, we focus on the problem of how CFs expand the attack surface of MLaaS by strengthening membership inference attacks (MIAs), and on the need to design defense mechanisms that mitigate this emerging risk without undermining utility and explainability. First, we systematically analyze how exposing CFs through query-based APIs enables more effective shadow-based MIAs. Second, we propose a defense framework that integrates Differential Privacy (DP) with Active Learning (AL) to jointly reduce memorization and limit effective training data exposure. Finally, we conduct an extensive empirical evaluation to characterize the three-way trade-off between privacy leakage, predictive performance, and explanation quality. Our findings highlight the need to carefully balance transparency, utility, and privacy in the responsible deployment of explainable MLaaS systems.

\end{abstract}
\begin{IEEEkeywords}
Counterfactual Explanations, Differential Privacy, Active Learning, Membership Inference Attack
\end{IEEEkeywords}

\section{Introduction}
Explainable AI (XAI) methods are increasingly employed in Machine Learning as a Service (MLaaS) systems with the aim of offering users and stakeholders more transparency into the decision-making process of deployed Machine learning (ML) models \cite{nguyen2025privacy,ezzeddine2024privacy,shokri2021privacy}. XAI frameworks provide insights into the factors underlying a model's prediction, allowing stakeholders, users, or regulators interpret automated decisions. For example, feature-importance methods attribute predictive influence to input variables, identifying which features drive a decision and how changes in their values affect model outputs \cite{montavon2019layer}. Counterfactual explanations (CFs) demonstrate how minimal input changes can alter a model's decision in a hypothetical scenario, highlighting which factors drive predictions \cite{wachter2017counterfactual}.

In standard MLaaS deployments, models are exposed via remote Application Programming Interfaces (APIs) that provide queriable prediction functionality. Through this query interface, users and stakeholders submit their data and receive the model’s outputs. While such API access is essential for MLaaS, it also creates a broad attack surface that can be exploited to compromise privacy by leveraging the model's responses. Several privacy attacks operate in this setting. Membership inference attacks (MIAs) aim to determine whether a specific record was included in the model's training set, model extraction attacks attempt to replicate the behavior of the target model, and property inference attacks seek to infer sensitive aggregate properties of the training data beyond the model's intended disclosures \cite{shokri2021privacy}. Integrating explanations into MLaaS, where, in addition to predictions, models also return explanations, introduces new security and privacy concerns. Recent research has further demonstrated that providing explanations alongside predictions can amplify these risks. This is because explanations, and particularly CFs, often reveal features, dependencies, and boundary regions on which the model relies, they provide adversaries with additional signals that can be systematically leveraged to infer sensitive information about the underlying training dataset, thereby increasing the effectiveness and precision of privacy attacks \cite{pawelczyk2023privacy,oksuz2024autolycus,wang2022dualcf,shokri2021privacy,ezzeddine2024privacy}.

Among these threats, MIAs are particularly critical because they directly undermine the confidentiality of training data. By revealing whether a specific individual's record was used to train a model, MIAs create a tangible link between the deployed model and identifiable individuals. Such leakage significantly increases the risk that the model itself may be considered personal data under the GDPR, since it can enable the re-identification of individuals \cite{veale2018algorithms}. This concern is especially acute when models are trained on highly sensitive datasets, where confirming membership may expose private attributes or sensitive participation. If explanations are available, attackers can further strengthen MIAs by combining standard model outputs (e.g., predicted probability distributions) with additional information disclosed through explanations. In particular, CFs can reveal how far an instance lies from the model's decision boundary and can be exploited by the intuition that training samples are more confidently classified and thus are often located farther from the decision boundary than non-members. Consequently, an adversary may infer membership when the distance between a given instance and its CF exceeds a pre-defined threshold \cite{pawelczyk2023privacy}.

In this context, it is essential to systematically analyze how CFs can be exploited to strengthen privacy attacks, such as MIAs, and to quantify the extent to which they enhance their effectiveness. At the same time, developing robust defense mechanisms is equally critical. Privacy-enhancing technologies naturally emerge as candidate solutions, and recent research has explored the application of Differential Privacy (DP), either to the underlying ML models or directly to the explanations \cite{ezzeddine2024knowledge,yang2022differentially,mochaourab2021robust}. However, introducing privacy-preserving mechanisms is not without cost. Applying DP may affect not only the model's predictive performance but also the usefulness of the generated explanations. Therefore, it is necessary to carefully evaluate how privacy protection impacts explanation utility, alongside model predictive performance. This tension gives rise to a fundamental interplay between three competing objectives: (i) resilience against privacy attacks such as MIAs, (ii) predictive performance of the underlying ML model, and (iii) the quality of the provided explanations.

\begin{figure*}[!ht]
    \centering
    \includegraphics[width=0.8\linewidth]{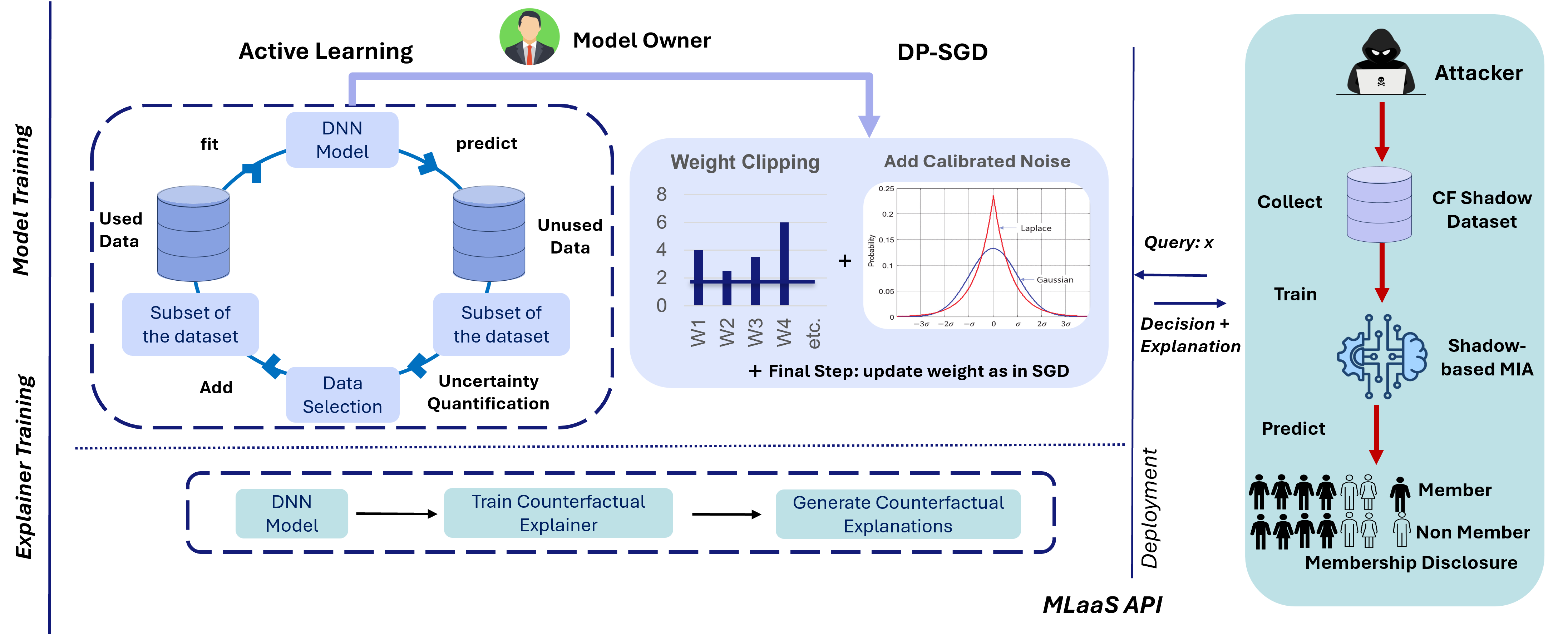}
    \caption{Scenario illustrating model training procedures performed, alongside the attacker's strategy of performing MIA.}
    \label{fig:scenario}
    \vspace{-0.6cm}
\end{figure*}

In this work, we address this three-way tension, with the broader goal of contributing to the responsible deployment of XAI in socially sensitive high-stakes domains such as finance and public services. It is imperative to ensure that efforts to enhance transparency do not expose individuals to privacy harms. To this end, we first investigate how CFs can be exploited to perform MIAs, in which an adversary has API query access to both predictions and CFs. In particular, we focus on \emph{shadow-based MIAs}, a practical, realistic, and well-established attack methodology, compared with \emph{threshold-based MIAs} \cite{sablayrolles2019white}, which rely on predefined thresholds of membership indicators such as loss or confidence. Specifically, in \emph{shadow-based MIAs}, attackers approximate the target model's behavior using surrogate models trained on auxiliary data. This assumption reflects realistic attack capabilities in deployment. Second, we propose a novel defense mechanism that integrates DP with Active Learning (AL)\footnote{Our intuition to use AL as a technique to further enhance privacy is discussed in more detail in Sec. \ref{Methodology}} as a data-distillation strategy \cite{munro2021human,sachdeva2023data,lei2023comprehensive,dong2022privacy,zheng2023differentially}. Our approach seeks to jointly combine DP and AL in a manner to strengthens resistance to MIAs while preserving both predictive performance and explanation quality (see Fig.~\ref{fig:scenario}). This design enables an evaluation of the trade-off between privacy preservation, measured by the attacker's ability to correctly infer membership, and explainability, quantified by CF quality.

To the best of our knowledge, this is among the first works to conduct a comprehensive study on shadow-based MIAs that leverage CFs and to propose a defense mechanism that combines DP and AL. Moreover, we systematically analyze the interplay between model predictive performance, attack success, and CF quality. Specifically, the contributions can be summarized as:
\begin{itemize}
    \item We systematically quantify how CFs can be leveraged as the adversary knowledge source for shadow-based MIA.
    \item We introduce a novel defense framework that integrates DP with AL to limit memorization and reduce training data exposure.
    \item We empirically characterize the three-way trade-off between privacy leakage, predictive performance, and CF quality, offering a unified analysis of privacy, utility, and explanation fidelity.
\end{itemize}

\section{Related Work}\label{relatedWork}

\emph{Explanation-based MIA} have emerged as powerful methods to perform threshold-based MIA \cite{ferry2025taming}. For instance, \cite{shokri2021privacy} demonstrates that back-propagation-based explanations strengthen MIA by leveraging the intuition that higher variance in feature attributions indicates greater uncertainty in the logits, signaling that an instance is less likely to be a training member. \cite{pawelczyk2023privacy} introduced an MIA recourse-based attack using CFs, exploiting the intuition that training points should be farther from decision boundaries than test points, predicting membership when the distance to a CF exceeds a threshold. Authors in \cite{liu2024please} further extend this understanding by introducing robustness-based MIA, where perturbations guided by attribution maps reveal that members exhibit confidence drops when important features are altered.

Other efforts have focused on defending against MIA \cite{yeom2018privacy,nasr2018machine}. In \cite{yeom2018privacy}, the authors propose regularization-based defenses and attribute the success of MIAs to overfitting. The authors of \cite{shokri2017membership} suggest overfitting-prevention techniques such as dropout and L2 regularization. Although helpful, these methods are not specifically designed for MIA defense and therefore lack robustness. Other methods have been designed specifically to defend against MIA. Authors in \cite{nasr2018machine} introduced an adversarial regularization method that adds a gain of membership inference term to the loss function, effectively creating a max-min game where the model maximizes and minimizes this term during training. \cite{li2021membership} proposed a regularizer between the softmax output distributions of the training and validation sets to prevent MIA. Another type of defense is adversarial example-based, which intentionally alters inputs or outputs, often through perturbations, to mislead an attacker. For instance, \cite{jia2019memguard} proposed adding crafted noise to the model's outputs, thereby transforming them into adversarial examples that mislead the attacker's classifier and prevent MIA.

Several other works have explored the use of DP to defend against MIA. \cite{abadi2016deep} proposed the DP-Stochastic Gradient Descent algorithm (DP-SGD), which builds on enhancing the original stochastic gradient descent by integrating DP into the backpropagation step. The work in \cite{rahman2018membership} revealed that even DP-SGD could still be compromised by MIA methods. To mitigate the drawbacks of employing DP during model training or data generation, other works have explored incorporating DP at different stages. For instance, \cite{ye2022one} modifies and normalizes confidence score vectors via a DP mechanism, thereby safeguarding privacy by obscuring membership information and reconstructed data, while \cite{yuan2022membership} provided an in-depth analysis of privacy risks in neural network pruning and discovered susceptibility to MIAs. In the interplay between XAI and privacy, the demonstrated tension between them has prompted research into methods that balance explainability with privacy guarantees \cite{naidu2021differential}. Studies reveal an inherent trade-off: applying privacy-preserving techniques such as DP to models and explainers often degrades the quality and utility of the explanations \cite{ezzeddine2024differential,ezzeddine2024knowledge}. In \cite{naidu2021differential}, it is demonstrated that significant visual degradation occurs in Deep Neural Network (DNNs) explanations when DP is applied \cite{saifullah2024privacy}. Despite these challenges, privacy-preserving xAI approaches have emerged. Authors in \cite{nori2021accuracy} successfully integrate DP into Explainable Boosting Machines, achieving promising accuracy with formal privacy guarantees, with the limitation of being model-specific.

In contrast to prior work, this is the first work to investigate how CFs can be exploited within a shadow‑based MIA framework, positioning CFs as a primary source of adversarial knowledge. Moreover, our work is the first to examine the combined use of DP and AL as a defense strategy against MIAs. While prior studies have primarily applied AL to reduce labeling costs or to generate synthetic datasets, we integrate AL with DP and assess how this combination strengthens privacy in MLaaS settings. The objective is to offer a comprehensive analysis of their interaction and its impact on model utility, membership leakage, and CF quality.

\section{Reference Scenario and Methodology}\label{Methodology}

\subsection{Reference Scenario: Shadow-based MIA}

Figure \ref{fig:scenario} illustrates our reference scenario. Consider a deep neural network (DNN) model \(f_{\theta}\) trained on a dataset \(\mathcal{D}\colon \mathbf{X} \in \mathbb{R}^d \to \mathbf{Y} \in \mathbb{R}^t\). The model \(f_{\theta}\) takes as input a numerical query \(\mathbf{x} \in \mathbf{X}\) and predicts an output \(\mathbf{y} \in \mathbf{Y}\). Let \(\theta\) denote the weight matrix and \(\mathbf{b}\) the bias vector of the DNN. The vector \(\mathbf{y} = f_{\theta}(\mathbf{x})\) represents confidence scores (probabilities), indicating the model's probability toward each class. The model \(f_{\theta}\) is deployed as an MLaaS, and the model owner trains a CF explainer \(E\). The CF explainer \(E\) is trained on the dataset \(\mathcal{D}\) to generate CF instances. The MLaaS returns \( { E(\mathbf{x}), f_{\theta}(\mathbf{x})}\ \) when queried by $x$. An attacker aims to determine whether specific data points belong to \(\mathcal{D}\) via MIA by exploiting the MLaaS API's output, as described in \cite{shokri2017membership}.

Specifically, the attacker collects a shadow dataset by repeatedly querying the API and distributes it into several datasets $\{\mathcal{D}_1, \mathcal{D}_2, \dots, \mathcal{D}_S\}$, each drawn from the same distribution $\mathcal{P}$ as $\mathcal{D}_{\mathrm{train}}$. For each dataset $\mathcal{D}_s$, the adversary trains a set of \emph{shadow models} $f_s(\cdot; \theta_s)$, aiming to approximate the behavior of the target model $f(\cdot; \theta)$. The adversary fully controls $\mathcal{D}_s$ and its membership status, so for every sample $x \in \mathcal{D}_s$, it is known that $x$ is a \emph{member}, while for $x \notin \mathcal{D}_s$ (but drawn from $\mathcal{P}$), it is a \emph{non-member}. Each shadow model $f_s$ returns a posterior vector $\boldsymbol{p}_s(x) = f_s(x; \theta_s) \in \mathbb{R}^K$, where $K$ is the number of classes. Using the shadow models posteriors \(\boldsymbol{p}_s(x)\) for both \emph{member} and \emph{non-member} samples, the adversary trains an \emph{attack model} $g(\cdot;\phi)$ to infer membership:
    \[
        g\bigl(\boldsymbol{p}_s(x);\phi\bigr) = 
        \begin{cases}
            1, & \text{if } x \in \mathcal{D}_s, \\
            0, & \text{otherwise}.
        \end{cases}
    \]

We evaluate the MIA across two configuration settings that differ in how the shadow dataset is constructed: i) No-CF setting and ii) CF-enabled setting. The collected dataset is then used to train the shadow models and indicates whether the MLaaS provider offers CF explanations. In both cases, the attacker must first assemble a shadow dataset that approximates the distribution of the original training data. Specifically:

\paragraph{No-CF setting}
The attacker does not have access to CF explanations, we follow the standard assumption in the literature: the shadow dataset contains model outputs for inputs drawn from the same distribution as the training data.
\[
\mathcal{D}_{\text{shadow}} = \{ f_{\theta}(\mathbf{x}_i) \}_{i=1}^{M},
\]

\paragraph{CF-enabled setting.}
The attacker leverages CF explanations provided by the MLaaS system, queries the CF explainer \(E\) for a set of input instances, and receives both the model outputs and CF explanations, thereby forming a shadow dataset. Access to CFs provides the attacker with additional side information \(E(x)\) that increases the mutual information between the model's outputs and the membership variable.
\[
\mathcal{D}_{\text{shadow}} = \{ E(\mathbf{x}_i),\, f_{\theta}(\mathbf{x}_i) \}_{i=1}^{M}.
\]
Therefore, the shadow model trained on $\mathcal{D}_{\text{shadow}}$ resulting in attack model \(g(p_s(x)\ \Vert\ \phi(E(x),x)\),
Note that while baselines such as CF-distance threshold exist, we do not include them because their performance depends strongly on the chosen CF generator and on feature scaling/normalization, making fair reproduction and threshold calibration ambiguous.

\subsection{Defense with Differential Privacy and Active Learning}
To defend against MIA, we formulate two main objectives (i) to limit the model's tendency to memorize training data during training, and (ii) to reduce the amount and sensitivity of data exposed. Therefore, we propose a defense mechanism where, during the model training phase, the model owner leverages both AL and DP. Specifically, we employ AL, despite being traditionally used to reduce annotation costs by selecting only the most informative samples, as a technique to train the DNN with the aim of reduce training data by constructing a smaller yet high-utility training data subset. Moreover, we employ DP, specifically, DP-SGD on the model level with rigorous privacy accounting to achieve a target $\varepsilon$-DP budget, to provide formal, instance-level privacy guarantees against MIAs. We refer to this approach as \emph{with DP}, i.e., where AL and DP are jointly employed, and consider an alternative approach, namely, \emph{without DP}, where DP is not employed. The goal is to evaluate the advantage of jointly leveraging AL and DP as a defense mechanism to mitigate CF-based shadow-based MIAs. Furthermore, we evaluate both model utility and the quality of the generated CFs, with the ultimate aim of analyzing how the proposed defense affects predictive performance and CF quality.

\emph{To initiate the AL process}, we construct a labeled subset \( \mathcal{S} \subset \mathcal{D} \) through random sampling. Initially, a small subset \( \mathcal{S}_0 \subset \mathcal{D} \) is randomly selected to serve as the starting point for labeling. At iteration \(t\), the current model \( f \) scores unlabeled data from \(\mathcal{D} \setminus \mathcal{S}\). An acquisition function (e.g., entropy or uncertainty sampling) measures how \emph{informative} each sample might be for improving the model. We select the most informative subset \(\mathcal{S}_t\) from the pool using the least confidence metric. After adding the samples in \(\mathcal{S}_t\), we incorporate them into the training set, i.e., \(\mathcal{S} \;\leftarrow\; \mathcal{S} \,\cup\, \mathcal{S}_t\). The updated training set is then used in the next model retraining step $t+1$. These steps are repeated until a stopping condition is met, or a maximum number of iterations on a validation set is reached. \emph{To integrate DP into the model training}, we update the model parameters following DP-SGD, where each weight in $\theta$ is updated during back-propagation step, where gradients computed using standard optimizers (e.g., SGD, Adam), are clipped to a fixed norm \(C\), and Gaussian noise is chosen to satisfy a target \((\epsilon,\delta)\)-DP guarantee. We consider the following scenarios:
\begin{itemize}
    \item \emph{Baseline}: Standard training without any additional techniques serving as a reference model for evaluating the benefits of DP and AL for preventing MIA.
    \item \emph{Only-AL}: Incorporates AL to strategically select training samples with the aim of reducing the required training dataset size while maintaining predictive performance.
    \item \emph{Only-DP}: Integrates DP into the training process of the model. It assesses the trade-offs between privacy and performance when DP is applied.
    \item \emph{DP-Post-AL}: Applies AL first to strategically select training samples, and DP is then subsequently applied during the training phase. This setup assesses whether applying DP after AL can improve privacy preservation.
    \item \emph{AL-Guided-DP}: Follows the same incremental training process as the default AL model, but with DP incorporated during each training step.
\end{itemize}

\section{Experimental Settings}\label{ExperimentalSettings}

\textbf{Datasets.} We consider two datasets for conducting our experiments, \emph{EEG} \cite{eegDataset} and \emph{Indoor Location} (In-Location) \cite{UjiIndoorLoc}. We selected these two datasets for their suitability for training DNN models in terms of performance and practical relevance, and for the variation in the dimensionality of their input features. The EEG dataset comprises 14,980 data records, 14 numerical features, and 2 classes. The dataset contains brainwave and signal readings, with the objective of training a model to predict whether an eye is open or closed. In-Location comprises 20,000 data records and 529 features related to Wi-Fi signals, along with multiple labels, and is used to predict the building.

\textbf{Model Architectures and Training Protocol.}
Preprocessing consists of removing missing values, removing outliers, and normalizing with a standard scaler. For both datasets, we split the data into 45\% for training the target model, 45\% for training shadow models, and 10\% for validation. For the EEG \emph{Baseline} model, we employed a DNN with 4 hidden layers containing 32, 16, 32, and 16 nodes, respectively. Each hidden layer was followed by a ReLU activation. The In-Location \emph{Baseline} model was constructed with 7 layers, with 600, 700, 600, 300, 600, 300, and 128 nodes, respectively, followed by ReLU activation function, and an additional dropout layer was used to prevent overfitting. These models were optimized using the Adam optimizer to minimize the cross-entropy loss, with a learning rate of 0.01.

\textbf{Defense Mechanisms.}
For scenarios where only DP was applied during training (i.e., \emph{Only-DP}), we maintained the same model architecture as the default. Models were trained with privacy budgets ($\varepsilon$) of 1, 3, 5, and 10, with lower budgets corresponding to stronger privacy guarantees, and with $\delta$ = 1e-5, grad normalization of 1.5. In the \emph{Only-AL} approach, the model during the first iteration is trained using 10\% of the available training set. With each iteration, an additional 50 data records are added; the process is repeated for 50 iterations, and the model achieving the highest accuracy is selected. The training subset used to train the optimal model is then saved as the \emph{Best AL Dataset}, on which the model achieves the best predictive performance. For model training in the \emph{DP-Post-AL}, we use the \emph{Best AL Dataset} and train models with DP. 

\textbf{Membership Inference Attack Setup.}
MIA is implemented by performing hyperparameter tuning on the number of shadow models (DNNs) and attack models (XGBoost) for the shadow models' outputs on the both datasets. After hyperparameter tuning, the XGB attack model is configured with 150 estimators, a maximum depth of 15, and a learning rate of 0.01. The shadow dataset is collected by using a separate subset of the dataset to query the MLaaS. The results are presented as the average of 5 runs.

\textbf{Counterfactual Explanation Setup.}
All CFs were obtained by using the Nearest Instance Counterfactual Explanations (NICE) \cite{niceAlgoArticle} with the following criteria: Distance Metric of Heterogeneous Euclidean-Overlap Metric, Normalization of Min-Max, and the algorithm was set to optimize proximity and sparsity when using the \emph{In-Location} dataset. CF quality was measured using well-known metrics, such as average proximity and sparsity. Each data point from the testing set was explained individually, and its proximity and sparsity values were used to compute the average proximity and sparsity across all data points. We gain insight into how effectively the CF explainer can produce \emph{concise} (i.e., small-distance, few-feature changes) yet \emph{impactful} (i.e., changing the predicted outcome) CFs under different privacy conditions.

\textbf{Evaluation Metrics Setup.}
Relative to MIA, we propose two distinct metrics. Let $\mathcal{D}$ denote the full dataset with $|\mathcal{D}| = N$ records. Let $\mathcal{S} \subseteq \mathcal{D}$ be the set of records used to train a given model (e.g., depending on whether AL is applied). Let $\hat{\mathcal{S}} \subseteq \mathcal{D}$ be the set of records that an adversary predicts as \emph{members} via MIA.
\paragraph*{Micro-level defended record ratio (per model)}
At the micro level, defended records quantify the fraction of \emph{true members} whose membership remains hidden, i.e., training points that the attacker \emph{fails} to infer as members, where \( \mathcal{D}_{\mathrm{protected}} \;=\; \mathcal{S} \setminus \hat{\mathcal{S}}\) and true positive rate (TPR) among all actual members, the fraction the attacker correctly detects.
\begin{equation}
    Micro=
    \frac{\lvert \mathcal{D}_{\mathrm{protected}} \rvert}{\lvert \mathcal{S} \rvert}
    = 1 - \mathrm{TPR}
\end{equation}

\paragraph*{Macro-level defended record ratio (overall exposure)}
At the macro level, we additionally account for records that were \emph{never} used for training and are therefore automatically unexposed to MIA. It describes the fraction of all records whose membership is not successfully revealed, i.e., records that are either saved (never trained) or protected despite being trained, where \(\mathcal{D}_{\mathrm{saved}} \;=\; \mathcal{D} \setminus \mathcal{S}\).
\begin{equation}
    Macro=
    \frac{\lvert \mathcal{D}_{\mathrm{saved}} \rvert + \lvert \mathcal{D}_{\mathrm{protected}} \rvert}{\lvert \mathcal{D} \rvert} 
    \;=\; 
    1-\frac{|\mathcal{S}|}{|\mathcal{D}|}\cdot \mathrm{Recall}_{\text{member}}.
\end{equation}

\section{Experimental Results}\label{Results}

\begin{table}[t]
\centering
\small
\setlength{\tabcolsep}{2.5pt}
\renewcommand{\arraystretch}{1.0}
\caption{Predictive performance (Accuracy, Precision, Recall) under privacy budgets $\varepsilon$ for EEG and In-Location. \textit{Baseline and Only-AL are trained without DP and therefore have no associated $\varepsilon$.}}
\begin{tabular}{clcccc|cccc|cccc}
\toprule
\multirow{3}{*}{Data} & \multirow{3}{*}{Method} 
& \multicolumn{4}{c|}{Accuracy} 
& \multicolumn{4}{c|}{Precision} 
& \multicolumn{4}{c}{Recall} \\

& & \multicolumn{12}{c}{$\varepsilon$} \\

\cmidrule(lr){3-6}\cmidrule(lr){7-10}\cmidrule(lr){11-14}
& & 1 & 3 & 5 & 10 
  & 1 & 3 & 5 & 10
  & 1 & 3 & 5 & 10 \\
\midrule

\multirow{5}{*}{\rotatebox{90}{EEG}}
& Baseline      
& \multicolumn{4}{c|}{96}
& \multicolumn{4}{c|}{96}
& \multicolumn{4}{c}{96} \\

& Only-AL       
& \multicolumn{4}{c|}{95}
& \multicolumn{4}{c|}{95}
& \multicolumn{4}{c}{95} \\

& Only-DP       
& 82 & 87 & 88 & 90
& 82 & 87 & 88 & 90
& 82 & 87 & 88 & 90 \\

& AL-Guided-DP  
& 82 & 87 & 90 & 91
& 82 & 88 & 90 & 92
& 82 & 87 & 90 & 91 \\

& DP-Post-AL    
& 66 & 72 & 83 & 86
& 69 & 74 & 84 & 86
& 66 & 72 & 83 & 86 \\

\midrule

\multirow{5}{*}{\rotatebox{90}{In-Location}}
& Baseline      
& \multicolumn{4}{c|}{100}
& \multicolumn{4}{c|}{100}
& \multicolumn{4}{c}{100} \\

& Only-AL       
& \multicolumn{4}{c|}{100}
& \multicolumn{4}{c|}{100}
& \multicolumn{4}{c}{100} \\

& Only-DP       
& 40 & 75 & 81 & 86
& 24 & 79 & 85 & 88
& 40 & 75 & 81 & 86 \\

& AL-Guided-DP  
& 58 & 57 & 56 & 91
& 46 & 56 & 67 & 92
& 58 & 57 & 56 & 91 \\

& DP-Post-AL    
& 47 & 45 & 52 & 53
& 23 & 29 & 50 & 54
& 47 & 45 & 52 & 53 \\

\bottomrule
\end{tabular}
\label{tab:Modelsperformance}
\end{table}

We present the numerical results along four complementary dimensions. First, we evaluate predictive performance under the different defense mechanisms. Second, we assess MIA effectiveness with and without CFs to quantify the additional privacy risk introduced by explanations and the robustness of the defenses. Third, we report micro- and macro-level defended record ratios to provide a granular view of privacy exposure. Finally, we analyze the impact of the defense strategies on CF quality.

\subsection{Model Predictive Performance}
Table~\ref{tab:Modelsperformance} summarizes the predictive performance of the different models across all approaches and both datasets.
Across both datasets, as expected, \emph{Baseline}, which does not implement any privacy measure, achieves the highest performance across all metrics (e.g., accuracy, precision, and recall of 96\% and 100\% across EEG and In-Location, respectively), while \emph{Only-AL}, which leverages only AL to reduce the training set size while ensuring performance matches that of \emph{Baseline}, closely follows (\emph{Only-AL} achieves this level of performance using only a subset of the dataset).  

When DP is incorporated as part of the defense mechanism, namely in \emph{Only-DP}, \emph{AL-Guided-DP}, and \emph{DP-Post-AL}, a consistent pattern emerges across both datasets. Stronger privacy guarantees (lower $\varepsilon$) lead to a noticeable degradation in performance, whereas performance steadily improves as $\varepsilon$ increases, corresponding to weaker privacy constraints. 

In the EEG dataset, \emph{AL-Guided-DP} and \emph{Only-DP} exhibit comparable performance and consistently outperform \emph{DP-Post-AL}, which yields the lowest scores. Specifically, \emph{AL-Guided-DP} achieves accuracy between 82\% and 91\%, with precision ranging from 82\% to 92\%, and recall following a similar pattern. \emph{Only-DP} attains accuracy between 82\% and 90\% across different $\varepsilon$ values. For In-Location, DP exerts a stronger impact on performance than in EEG, although the overall trend remains similar. \emph{Only-DP} and \emph{AL-Guided-DP} alternate in achieving the best predictive results, with comparable performance levels. Importantly, \emph{AL-Guided-DP} attains this performance while relying on fewer training samples (since it employs AL), which is beneficial from a privacy perspective as it reduces potential exposure to MIAs. 

Regarding the subset size selected by AL, for EEG, \emph{Only-AL} (and \emph{DP-Post-AL}) uses 77.45\% of the data. \emph{AL-Guided-DP} selects 78.97\%, 74.48\%, 56.49\%, and 66.98\% for $\varepsilon = 1, 3, 5,$ and $10$, respectively. For In-Location, \emph{Only-AL} uses only 10\% of the data, while \emph{DP-Post-AL} requires 29.13\%, 21\%, and 40.35\% for $\varepsilon = 1, 3,$ and $5$. Overall, AL can substantially reduce the required training data; however, when combined with DP, larger subsets are often selected, reflecting the additional uncertainty introduced by DP noise.

\emph{Takeaway 1:} AL preserves utility, matching baseline performance while using fewer training samples. Introducing DP leads to a utility loss at low $\varepsilon$. While \emph{AL-Guided-DP} does not substantially outperform \emph{Only-DP}, it achieves comparable performance with fewer samples. In contrast, \emph{DP-Post-AL} consistently yields the lowest performance across both datasets.

\subsection{Membership Inference Attack}
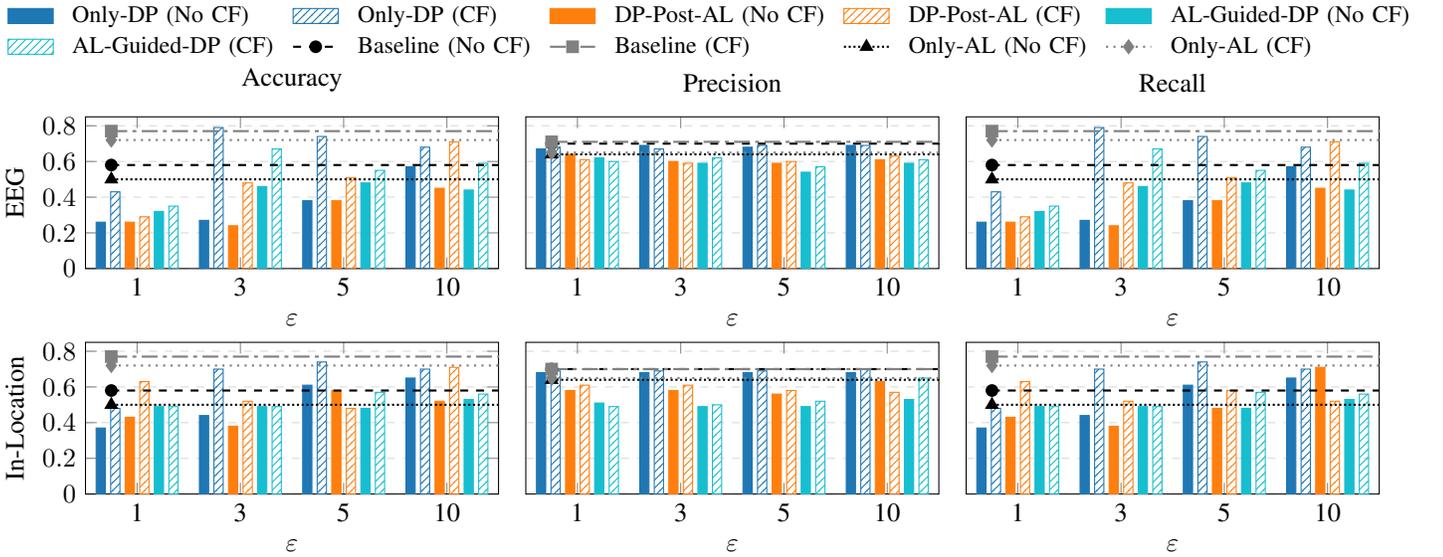
\begin{figure*}[!hbt]
\centering
\begin{tikzpicture}
\begin{axis}[
  hide axis,
  xmin=0, xmax=1, ymin=0, ymax=1,
  legend columns=5,
  legend style={
    font=\small,
    draw=none, fill=none,
    inner sep=1pt,
    row sep=0pt,
    column sep=5pt,
    legend cell align=left,
  },
]
  \addlegendimage{ybar, area legend, draw=myblue, fill=myblue}
  \addlegendentry{Only-DP (No CF)}
  \addlegendimage{ybar, area legend, draw=myblue, fill=myblue,
                  pattern=north east lines, pattern color=myblue}
  \addlegendentry{Only-DP (CF)}

  \addlegendimage{ybar, area legend, draw=myorange, fill=myorange}
  \addlegendentry{DP-Post-AL (No CF)}
  \addlegendimage{ybar, area legend, draw=myorange, fill=myorange,
                  pattern=north east lines, pattern color=myorange}
  \addlegendentry{DP-Post-AL (CF)}

  \addlegendimage{ybar, area legend, draw=myteal, fill=myteal}
  \addlegendentry{AL-Guided-DP (No CF)}
  \addlegendimage{ybar, area legend, draw=myteal, fill=myteal,
                  pattern=north east lines, pattern color=myteal}
  \addlegendentry{AL-Guided-DP (CF)}

  \addlegendimage{black, thick, dashed,
                  mark=*, mark options={solid, fill=black}, mark size=2pt}
  \addlegendentry{Baseline (No CF)}
  \addlegendimage{gray, thick,
                  dash pattern=on 6pt off 2pt on 1.5pt off 2pt,
                  mark=square*, mark options={solid, fill=gray}, mark size=2pt}
  \addlegendentry{Baseline (CF)}
  \addlegendimage{black, thick, densely dotted,
                  mark=triangle*, mark options={solid, fill=black}, mark size=2.3pt}
  \addlegendentry{Only-AL (No CF)}
  \addlegendimage{gray, thick, dotted,
                  mark=diamond*, mark options={solid, fill=gray}, mark size=2.3pt}
  \addlegendentry{Only-AL (CF)}
\end{axis}
\end{tikzpicture}

\vspace{-2pt}

\begin{tikzpicture}
\pgfplotsset{
  miaAxis/.style={
    width=0.39\textwidth,
    height=3.6cm,
    xmin=-1, xmax=7,
    xtick={0,2,4,6},
    xticklabels={1,3,5,10},
    xlabel={$\varepsilon$},
    ymajorgrids,
    grid style={dashed, gray!30},
  },
  miaBars/.style={
    ybar,
    bar width=3.5pt,
    bar shift auto,
    mark=none,
    area legend,
  },
  miaLine/.style={mark=none}
}

\hspace{-3pt}

\begin{groupplot}[
  miaAxis,
  group style={group size=3 by 1, horizontal sep=0.02\textwidth},
]

\nextgroupplot[title={Accuracy}, ylabel={EEG}, ymin=0, ymax=0.85]
\addplot[miaBars, draw=myblue, fill=myblue]
  coordinates {(0,0.26) (2,0.27) (4,0.38) (6,0.57)};
\addplot[miaBars, draw=myblue, fill=myblue, pattern=north east lines, pattern color=myblue]
  coordinates {(0,0.43) (2,0.79) (4,0.74) (6,0.68)};
\addplot[miaBars, draw=myorange, fill=myorange]
  coordinates {(0,0.26) (2,0.24) (4,0.38) (6,0.45)};
\addplot[miaBars, draw=myorange, fill=myorange, pattern=north east lines, pattern color=myorange]
  coordinates {(0,0.29) (2,0.48) (4,0.51) (6,0.71)};
\addplot[miaBars, draw=myteal, fill=myteal]
  coordinates {(0,0.32) (2,0.46) (4,0.48) (6,0.44)};
\addplot[miaBars, draw=myteal, fill=myteal, pattern=north east lines, pattern color=myteal]
  coordinates {(0,0.35) (2,0.67) (4,0.55) (6,0.59)};

\addplot[miaLine, black, thick, dashed,
         mark=*, mark options={solid, fill=black}, mark size=2pt, mark repeat=2]
  coordinates {(-0.5,0.58) (7.5,0.58)};
\addplot[miaLine, gray, thick,
         dash pattern=on 6pt off 2pt on 1.5pt off 2pt,
         mark=square*, mark options={solid, fill=gray}, mark size=2pt, mark repeat=2]
  coordinates {(-0.5,0.77) (7.5,0.77)};
\addplot[miaLine, black, thick, densely dotted,
         mark=triangle*, mark options={solid, fill=black}, mark size=2.3pt, mark repeat=2]
  coordinates {(-0.5,0.50) (7.5,0.50)};
\addplot[miaLine, gray, thick, dotted,
         mark=diamond*, mark options={solid, fill=gray}, mark size=2.3pt, mark repeat=2]
  coordinates {(-0.5,0.72) (7.5,0.72)};

\nextgroupplot[
  title={Precision}, ymin=0, ymax=0.85,
  yticklabels=\empty, y tick style={draw=none}, ylabel={}
]
\addplot[miaBars, draw=myblue, fill=myblue]
  coordinates {(0,0.67) (2,0.69) (4,0.68) (6,0.69)};
\addplot[miaBars, draw=myblue, fill=myblue, pattern=north east lines, pattern color=myblue]
  coordinates {(0,0.68) (2,0.67) (4,0.69) (6,0.69)};
\addplot[miaBars, draw=myorange, fill=myorange]
  coordinates {(0,0.64) (2,0.60) (4,0.59) (6,0.61)};
\addplot[miaBars, draw=myorange, fill=myorange, pattern=north east lines, pattern color=myorange]
  coordinates {(0,0.61) (2,0.59) (4,0.60) (6,0.63)};
\addplot[miaBars, draw=myteal, fill=myteal]
  coordinates {(0,0.62) (2,0.59) (4,0.54) (6,0.59)};
\addplot[miaBars, draw=myteal, fill=myteal, pattern=north east lines, pattern color=myteal]
  coordinates {(0,0.60) (2,0.62) (4,0.57) (6,0.61)};

\addplot[miaLine, black, thick, dashed,
         mark=*, mark options={solid, fill=black}, mark size=2pt, mark repeat=2]
  coordinates {(-0.5,0.70) (7.5,0.70)};
\addplot[miaLine, gray, thick,
         dash pattern=on 6pt off 2pt on 1.5pt off 2pt,
         mark=square*, mark options={solid, fill=gray}, mark size=2pt, mark repeat=2]
  coordinates {(-0.5,0.71) (7.5,0.71)};
\addplot[miaLine, black, thick, densely dotted,
         mark=triangle*, mark options={solid, fill=black}, mark size=2.3pt, mark repeat=2]
  coordinates {(-0.5,0.64) (7.5,0.64)};
\addplot[miaLine, gray, thick, dotted,
         mark=diamond*, mark options={solid, fill=gray}, mark size=2.3pt, mark repeat=2]
  coordinates {(-0.5,0.65) (7.5,0.65)};

\nextgroupplot[
  title={Recall}, ymin=0, ymax=0.85,
  yticklabels=\empty, y tick style={draw=none}, ylabel={}
]
\addplot[miaBars, draw=myblue, fill=myblue]
  coordinates {(0,0.26) (2,0.27) (4,0.38) (6,0.57)};
\addplot[miaBars, draw=myblue, fill=myblue, pattern=north east lines, pattern color=myblue]
  coordinates {(0,0.43) (2,0.79) (4,0.74) (6,0.68)};
\addplot[miaBars, draw=myorange, fill=myorange]
  coordinates {(0,0.26) (2,0.24) (4,0.38) (6,0.45)};
\addplot[miaBars, draw=myorange, fill=myorange, pattern=north east lines, pattern color=myorange]
  coordinates {(0,0.29) (2,0.48) (4,0.51) (6,0.71)};
\addplot[miaBars, draw=myteal, fill=myteal]
  coordinates {(0,0.32) (2,0.46) (4,0.48) (6,0.44)};
\addplot[miaBars, draw=myteal, fill=myteal, pattern=north east lines, pattern color=myteal]
  coordinates {(0,0.35) (2,0.67) (4,0.55) (6,0.59)};

\addplot[miaLine, black, thick, dashed,
         mark=*, mark options={solid, fill=black}, mark size=2pt, mark repeat=2]
  coordinates {(-0.5,0.58) (7.5,0.58)};
\addplot[miaLine, gray, thick,
         dash pattern=on 6pt off 2pt on 1.5pt off 2pt,
         mark=square*, mark options={solid, fill=gray}, mark size=2pt, mark repeat=2]
  coordinates {(-0.5,0.77) (7.5,0.77)};
\addplot[miaLine, black, thick, densely dotted,
         mark=triangle*, mark options={solid, fill=black}, mark size=2.3pt, mark repeat=2]
  coordinates {(-0.5,0.50) (7.5,0.50)};
\addplot[miaLine, gray, thick, dotted,
         mark=diamond*, mark options={solid, fill=gray}, mark size=2.3pt, mark repeat=2]
  coordinates {(-0.5,0.72) (7.5,0.72)};

\end{groupplot}

\begin{scope}[yshift=-3cm]
\begin{groupplot}[
  miaAxis,
  group style={group size=3 by 1, horizontal sep=0.02\textwidth},
]

\nextgroupplot[ylabel={In-Location}, ymin=0, ymax=0.85]
\addplot[miaBars, draw=myblue, fill=myblue]
  coordinates {(0,0.37) (2,0.44) (4,0.61) (6,0.65)};
\addplot[miaBars, draw=myblue, fill=myblue, pattern=north east lines, pattern color=myblue]
  coordinates {(0,0.48) (2,0.70) (4,0.74) (6,0.70)};
\addplot[miaBars, draw=myorange, fill=myorange]
  coordinates {(0,0.43) (2,0.38) (4,0.58) (6,0.52)};
\addplot[miaBars, draw=myorange, fill=myorange, pattern=north east lines, pattern color=myorange]
  coordinates {(0,0.63) (2,0.52) (4,0.48) (6,0.71)};
\addplot[miaBars, draw=myteal, fill=myteal]
  coordinates {(0,0.49) (2,0.49) (4,0.48) (6,0.53)};
\addplot[miaBars, draw=myteal, fill=myteal, pattern=north east lines, pattern color=myteal]
  coordinates {(0,0.49) (2,0.49) (4,0.57) (6,0.56)};

\addplot[miaLine, black, thick, dashed,
         mark=*, mark options={solid, fill=black}, mark size=2pt, mark repeat=2]
  coordinates {(-0.5,0.58) (7.5,0.58)};
\addplot[miaLine, gray, thick,
         dash pattern=on 6pt off 2pt on 1.5pt off 2pt,
         mark=square*, mark options={solid, fill=gray}, mark size=2pt, mark repeat=2]
  coordinates {(-0.5,0.77) (7.5,0.77)};
\addplot[miaLine, black, thick, densely dotted,
         mark=triangle*, mark options={solid, fill=black}, mark size=2.3pt, mark repeat=2]
  coordinates {(-0.5,0.50) (7.5,0.50)};
\addplot[miaLine, gray, thick, dotted,
         mark=diamond*, mark options={solid, fill=gray}, mark size=2.3pt, mark repeat=2]
  coordinates {(-0.5,0.72) (7.5,0.72)};

\nextgroupplot[ymin=0, ymax=0.85, yticklabels=\empty, y tick style={draw=none}, ylabel={}]
\addplot[miaBars, draw=myblue, fill=myblue]
  coordinates {(0,0.68) (2,0.68) (4,0.68) (6,0.68)};
\addplot[miaBars, draw=myblue, fill=myblue, pattern=north east lines, pattern color=myblue]
  coordinates {(0,0.69) (2,0.69) (4,0.69) (6,0.70)};
\addplot[miaBars, draw=myorange, fill=myorange]
  coordinates {(0,0.58) (2,0.58) (4,0.56) (6,0.63)};
\addplot[miaBars, draw=myorange, fill=myorange, pattern=north east lines, pattern color=myorange]
  coordinates {(0,0.61) (2,0.61) (4,0.58) (6,0.57)};
\addplot[miaBars, draw=myteal, fill=myteal]
  coordinates {(0,0.51) (2,0.49) (4,0.49) (6,0.53)};
\addplot[miaBars, draw=myteal, fill=myteal, pattern=north east lines, pattern color=myteal]
  coordinates {(0,0.49) (2,0.50) (4,0.52) (6,0.65)};

\addplot[miaLine, black, thick, dashed,
         mark=*, mark options={solid, fill=black}, mark size=2pt, mark repeat=2]
  coordinates {(-0.5,0.70) (7.5,0.70)};
\addplot[miaLine, gray, thick,
         dash pattern=on 6pt off 2pt on 1.5pt off 2pt,
         mark=square*, mark options={solid, fill=gray}, mark size=2pt, mark repeat=2]
  coordinates {(-0.5,0.70) (7.5,0.70)};
\addplot[miaLine, black, thick, densely dotted,
         mark=triangle*, mark options={solid, fill=black}, mark size=2.3pt, mark repeat=2]
  coordinates {(-0.5,0.64) (7.5,0.64)};
\addplot[miaLine, gray, thick, dotted,
         mark=diamond*, mark options={solid, fill=gray}, mark size=2.3pt, mark repeat=2]
  coordinates {(-0.5,0.65) (7.5,0.65)};

\nextgroupplot[ymin=0, ymax=0.85, yticklabels=\empty, y tick style={draw=none}, ylabel={}]
\addplot[miaBars, draw=myblue, fill=myblue]
  coordinates {(0,0.37) (2,0.44) (4,0.61) (6,0.65)};
\addplot[miaBars, draw=myblue, fill=myblue, pattern=north east lines, pattern color=myblue]
  coordinates {(0,0.48) (2,0.70) (4,0.74) (6,0.70)};
\addplot[miaBars, draw=myorange, fill=myorange]
  coordinates {(0,0.43) (2,0.38) (4,0.48) (6,0.71)};
\addplot[miaBars, draw=myorange, fill=myorange, pattern=north east lines, pattern color=myorange]
  coordinates {(0,0.63) (2,0.52) (4,0.58) (6,0.52)};
\addplot[miaBars, draw=myteal, fill=myteal]
  coordinates {(0,0.49) (2,0.49) (4,0.48) (6,0.53)};
\addplot[miaBars, draw=myteal, fill=myteal, pattern=north east lines, pattern color=myteal]
  coordinates {(0,0.49) (2,0.49) (4,0.57) (6,0.56)};

\addplot[miaLine, black, thick, dashed,
         mark=*, mark options={solid, fill=black}, mark size=2pt, mark repeat=2]
  coordinates {(-0.5,0.58) (7.5,0.58)};
\addplot[miaLine, gray, thick,
         dash pattern=on 6pt off 2pt on 1.5pt off 2pt,
         mark=square*, mark options={solid, fill=gray}, mark size=2pt, mark repeat=2]
  coordinates {(-0.5,0.77) (7.5,0.77)};
\addplot[miaLine, black, thick, densely dotted,
         mark=triangle*, mark options={solid, fill=black}, mark size=2.3pt, mark repeat=2]
  coordinates {(-0.5,0.50) (7.5,0.50)};
\addplot[miaLine, gray, thick, dotted,
         mark=diamond*, mark options={solid, fill=gray}, mark size=2.3pt, mark repeat=2]
  coordinates {(-0.5,0.72) (7.5,0.72)};
\end{groupplot}
\end{scope}
\end{tikzpicture}
\caption{\footnotesize{Accuracy, precision, and recall across baselines, privacy budgets $\varepsilon$ with and without using CFs for MIA.}}
\label{fig:MIA-rows-bars-lines-toplegend-tight}
\vspace{-0.6cm}
\end{figure*}

Figure~\ref{fig:MIA-rows-bars-lines-toplegend-tight} summarizes MIA performance across all approaches and privacy budgets $\varepsilon$ of 1,3,5 and 10 (lower values indicate stronger privacy protection). Results show a clear trend across all cases: \emph{leveraging CFs to perform MIA allows for substantially amplifying attack success}. For instance, in the EEG dataset, \emph{Baseline (No CF)} attains $58\%$ accuracy/recall and $70\%$ precision while \emph{Baseline (CF)} attains $77\%$ of accuracy/recall, and $71\%$ of precision. Another example is \emph{Only-AL (No CF)}, accuracy/recall increases from $50\%$ to $72\%$ (precision $64$-$65\%$) when leveraging CFs (\emph{Only AL (CF)}. When employing DP, the impact of CFs on MIA success becomes even more evident. Comparing \emph{Only-DP (no CF)} to \emph{Only-DP (CF)}, accuracy/recall increases from $26\%$ to $43\%$ at $\varepsilon = 1$, from $27\%$ to $79\%$ at $\varepsilon = 3$, from $38\%$ to $74\%$ at $\varepsilon = 5$, and from $57\%$ to $68\%$ at $\varepsilon = 10$ when CFs are leveraged, while precision remains essentially close ($67$--$69\%$). Results in In-Location show the same trend: leveraging CFs consistently strengthens MIA. For example, \emph{Only-DP}, leveraging CFs raises accuracy/recall from $37,44,61,65\%$ to $48,70,74,70\%$ at $\varepsilon=1,3,5,10$. For \emph{DP-Post-AL}, accuracy increases from $43,38,58,52\%$ to $63,52,48,71\%$, except at $\varepsilon=5$. \emph{AL-Guided-DP} shows little change at $\varepsilon=1,3$, a moderate gain at $\varepsilon=5$, and at $\varepsilon=10$ its largest precision increase ($53\%$ to $65\%$) with a slight accuracy rise ($53\%$ to $56\%$). 
\emph{Takeaway 2:} While DP introduces a utility–privacy trade-off (Takeaway~1), leveraging CFs significantly weakens privacy protection across all settings a consistently amplifies MIA performance, often dramatically increasing attack success, even under DP. In several cases, CFs nearly double accuracy/recall at moderate privacy budgets, demonstrating that explanations can substantially undermine the privacy gains achieved by DP alone.


We now quantify the impact of $\varepsilon$ on the defense against MIA. As expected, a larger $\varepsilon$ weakens privacy: for example, on EEG, \emph{Only-DP (No CF)} accuracy/recall range from $26\%$ to $57\%$ and from $43\%$ to $68\%$ \emph{Only-DP (CF)}. Comparable monotonic trends appear in \emph{DP-Post-AL} and \emph{AL-Guided-DP}. We move to comparing \emph{Only-AL to DP-Post-AL} in order to better understand the effect of employing DP after AL. For EEG, \emph{Only-AL (No CF} has an accuracy/recall $\approx 50\%$, which reaches $72\%$ when CFs are exposed. In contrast, \emph{DP-Post-AL} starts from lower attack success at small privacy budgets ($26\%$ and $34\%$ at $\varepsilon$ 1 and 3), indicating stronger protection than \emph{Only-AL}. However, CFs significantly erode this advantage: accuracy rises to $29\%$ and $48\%$ at $\varepsilon=1$ and $3$, and at larger $\varepsilon$ of 10 reaching $71\%$. For In-Location, Similar trends are observed.


\emph{Takeaway 3:} Consistent with the utility–privacy trade-off observed in Takeaway~1, increasing $\varepsilon$ systematically weakens protection and enables stronger MIAs. However, beyond this expected trend, releasing CFs introduces an additional attack surface that substantially boosts member detection, primarily by increasing recall. Even when DP reduces attack success at low privacy budgets, this advantage is significantly diminished once CFs are exposed. While combining AL with DP can partially mitigate attack success, this comes at the cost of retaining more training data.

\begin{figure*}[!ht]
\centering
\begin{tikzpicture}
\begin{groupplot}[
  group style={
    group size=4 by 2,
    horizontal sep=1.3cm,
    vertical sep=1.3cm,
  },
  width=4.5cm,
  height=3cm,
  xlabel={$\varepsilon$},
  xlabel style={yshift=10pt},
  ylabel={\scriptsize\% def.\ records},
  xtick={1,3,5,10},
  ymin=0, ymax=110,
  grid=both,
  clip=false,
]

\nextgroupplot[
  legend to name=commonlegend,
  legend columns=2,
  legend style={
    font=\small,
    draw=none,
    fill=none,
    /tikz/every even column/.append style={column sep=10pt},
  },
]
\addplot+[mark=*, blue, thick] coordinates {(1,97) (3,100) (5,68) (10,56)};
\addlegendentry{MIA Without CF}
\addplot+[mark=square*, red, thick] coordinates {(1,90) (3,48) (5,44) (10,9)};
\addlegendentry{MIA with CF}
\node at (rel axis cs:0.5,-0.48) {\footnotesize (a) DP-Post-AL Micro};

\nextgroupplot
\addplot+[mark=*, blue, thick] coordinates {(1,97.70) (3,100) (5,75.24) (10,66)};
\addplot+[mark=square*, red, thick] coordinates {(1,92.28) (3,59.74) (5,56.64) (10,29.53)};
\node at (rel axis cs:0.5,-0.48) {\footnotesize (b) DP-Post-AL Macro};

\nextgroupplot
\addplot+[mark=*, blue, thick] coordinates {(1,85) (3,54) (5,55) (10,65)};
\addplot+[mark=square*, red, thick] coordinates {(1,77) (3,14) (5,32) (10,28)};
\node at (rel axis cs:0.5,-0.48) {\footnotesize (c) AL-Guided-DP Micro};

\nextgroupplot
\addplot+[mark=*, blue, thick] coordinates {(1,88.15) (3,65.74) (5,74.58) (10,76.56)};
\addplot+[mark=square*, red, thick] coordinates {(1,81.84) (3,35.95) (5,61.59) (10,51.78)};
\node at (rel axis cs:0.5,-0.48) {\footnotesize (d) AL-Guided-DP Macro};

\nextgroupplot
\addplot+[mark=*, blue, thick] coordinates {(1,74) (3,47) (5,71) (10,65)};
\addplot+[mark=square*, red, thick] coordinates {(1,39) (3,26) (5,57) (10,47)};
\node at (rel axis cs:0.5,-0.48) {\footnotesize (e) DP-Post-AL Micro};

\nextgroupplot[ymin=80, ymax=100]
\addplot+[mark=*, blue, thick] coordinates {(1,97.40) (3,94.70) (5,97.10) (10,96.50)};
\addplot+[mark=square*, red, thick] coordinates {(1,93.90) (3,92.60) (5,95.70) (10,94.70)};
\node at (rel axis cs:0.5,-0.48) {\footnotesize (f) DP-Post-AL Macro};

\nextgroupplot
\addplot+[mark=*, blue, thick] coordinates {(1,72) (3,58) (5,56) (10,39)};
\addplot+[mark=square*, red, thick] coordinates {(1,51) (3,17) (5,44) (10,30)};
\node at (rel axis cs:0.5,-0.48) {\footnotesize (g) AL-Guided-DP Micro};

\nextgroupplot
\addplot+[mark=*, blue, thick] coordinates {(1,91.84) (3,89.98) (5,90.37) (10,75.38)};
\addplot+[mark=square*, red, thick] coordinates {(1,85.72) (3,80.20) (5,71.75) (10,51.78)};
\node at (rel axis cs:0.5,-0.48) {\footnotesize (h) AL-Guided-DP Macro};

\end{groupplot}

\node[anchor=south] at ($(current bounding box.north)+(0mm,0)$)
  {\pgfplotslegendfromname{commonlegend}};

\end{tikzpicture}

\caption{\footnotesize{
Percentage of micro \& macro defended records under \emph{DP-Post-AL} \& \emph{AL-Guided-DP}, with \& without CFs. Top row: EEG; bottom row: In-Location.
}}
\label{fig:MicroMacro-EEG-InLocation}
\vspace{-0.3cm}
\end{figure*}
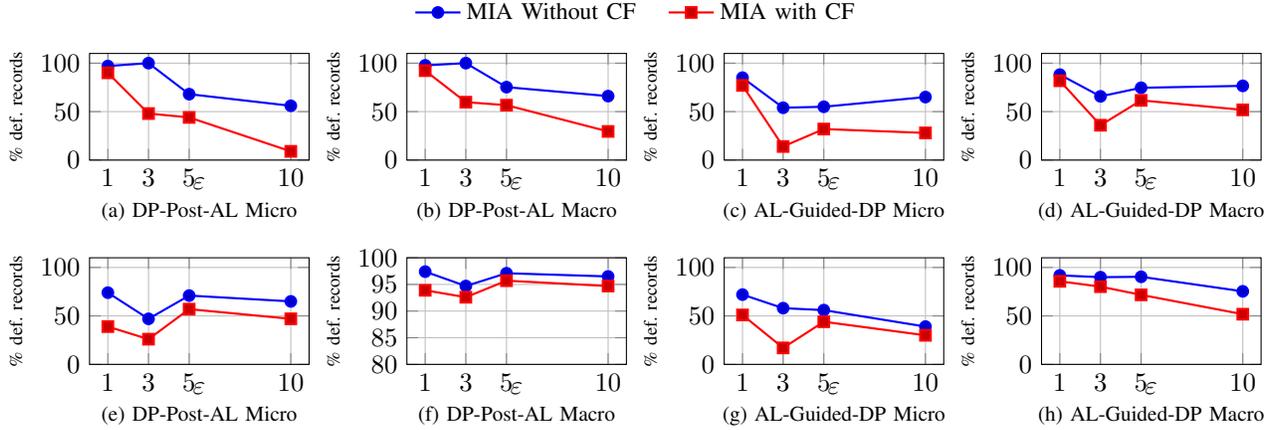

\begin{figure*}[!ht]
\centering
\begin{tikzpicture}
\begin{groupplot}[
  group style={
    group size=4 by 1,
    horizontal sep=1.6cm,
  },
  width=4.5cm,
  height=3.4cm,
  grid=both,
  xtick={1,3,5,7},
  xticklabels={1,3,5,10},
  xlabel={\(\varepsilon\)},
  tick label style={font=\small},
  label style={font=\small},
  title style={font=\footnotesize, align=center},
]

\nextgroupplot[
  title={EEG Avg Proximity},
  ylabel={Proximity},
  legend to name=CFlegend,
  legend columns=5,
  legend style={
    font=\small,
    draw=none,
    fill=none,
    inner sep=1pt,
    /tikz/every even column/.append style={column sep=8pt},
  },
]
\addplot+[mark=*, thick, color=blue]   coordinates {(1,5.7813) (3,5.9708) (5,6.2226) (7,5.9777)};
\addlegendentry{Only-DP}
\addplot+[mark=triangle*, thick, color=orange] coordinates {(1,7.1525) (3,5.6049) (5,6.2182) (7,6.1133)};
\addlegendentry{DP-Post-AL}
\addplot+[mark=square*, thick, color=teal] coordinates {(1,6.2866) (3,6.4253) (5,6.0872) (7,6.2305)};
\addlegendentry{AL-Guided-DP}
\addplot+[dashed, black, thick] coordinates {(1,6.7907) (7,6.7907)};
\addlegendentry{Baseline}
\addplot+[dotted, black, thick] coordinates {(1,6.6564) (7,6.6564)};
\addlegendentry{Only-AL}

\nextgroupplot[
  title={EEG Avg Sparsity},
  ylabel={Sparsity},
]
\addplot+[mark=*, thick, color=blue]   coordinates {(1,0.5017) (3,0.4559) (5,0.4520) (7,0.4391)};
\addplot+[mark=triangle*, thick, color=orange] coordinates {(1,0.6108) (3,0.4523) (5,0.4940) (7,0.4542)};
\addplot+[mark=square*, thick, color=teal] coordinates {(1,0.4986) (3,0.4527) (5,0.4515) (7,0.4491)};
\addplot+[dashed, black, thick] coordinates {(1,0.4712) (7,0.4712)};
\addplot+[dotted, black, thick] coordinates {(1,0.4834) (7,0.4834)};
\hspace{10pt}
\nextgroupplot[
  title={In-Location Avg Proximity},
  ylabel={Proximity},
  ymin=0, ymax=2800,
]
\addplot+[mark=*, thick, color=blue]   coordinates {(1,300.09) (3,393.62) (5,657.65) (7,697.63)};
\addplot+[mark=triangle*, thick, color=orange] coordinates {(1,269.57) (3,319.82) (5,367.64) (7,316.10)};
\addplot+[mark=square*, thick, color=teal] coordinates {(1,272.06) (3,285.69) (5,290.30) (7,332.92)};
\addplot+[dashed, black, thick] coordinates {(1,1410.33) (7,1410.33)};
\addplot+[dotted, black, thick] coordinates {(1,2573.60) (7,2573.60)};

\nextgroupplot[
  title={In-Location Avg Sparsity},
  ylabel={Sparsity},
]
\addplot+[mark=*, thick, color=blue]   coordinates {(1,1.76) (3,2.22) (5,3.71) (7,4.13)};
\addplot+[mark=triangle*, thick, color=orange] coordinates {(1,1.57) (3,1.82) (5,2.08) (7,1.78)};
\addplot+[mark=square*, thick, color=teal] coordinates {(1,1.63) (3,1.65) (5,1.68) (7,1.88)};
\addplot+[dashed, black, thick] coordinates {(1,7.94) (7,7.94)};
\addplot+[dotted, black, thick] coordinates {(1,14.39) (7,14.39)};

\end{groupplot}

\node[anchor=south] at ($ (group c2r1.north)!0.5!(group c3r1.north) + (0,6mm) $)
{\pgfplotslegendfromname{CFlegend}};

\end{tikzpicture}
\caption{\footnotesize{Sparsity and proximity of extracted CFs for EEG and In-Location across differential privacy budgets ($\varepsilon$).}}
\label{Combined_CF}
\vspace{-0.6cm}
\end{figure*}
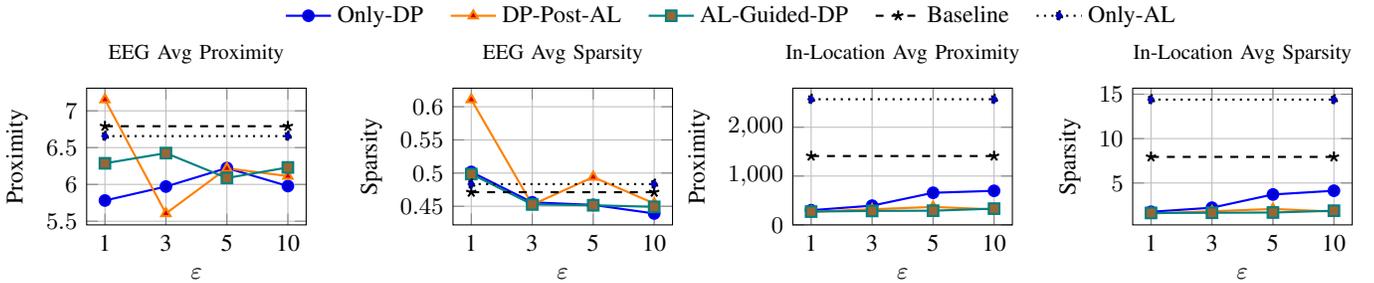

\subsection{Defense at Micro- and Macro-Level}

In this part we focus our analysis on the two main approaches, \emph{DP-Post-AL} and \emph{AL-Guided-DP}, and examine their behavior through the micro and macro defended record ratios. The micro metric captures the fraction of \emph{true training records} whose membership remains hidden, directly quantifying how well individual members are protected. In contrast, the macro metric reflects overall dataset exposure by additionally accounting for records that were never used for training and are therefore automatically unexposed to MIA. Consequently, macro values can remain high even when the protection of actual training points deteriorates.

Across all settings, leveraging CFs consistently reduces the proportion of defended points at both levels, with the effect being particularly pronounced at the micro level. On EEG, under \emph{DP-Post-AL}, micro protection drops from $68\%$ to $44\%$ at $\varepsilon=5$, and from $56\%$ to $9\%$ at $\varepsilon=10$ when CFs are leveraged. A similar trend is observed for \emph{AL-Guided-DP}, where micro defended rates decrease from $54\%$ to $14\%$ at $\varepsilon=3$. These results indicate that CFs substantially increase correct member inference, directly exposing training data.

At the macro level, defended ratios also decrease when CFs are used, but the reduction is often less because macro protection includes automatically saved (non-trained) records. As a result, an approach may appear robust at the macro level while failing to adequately protect actual training members. For instance, in EEG at $\varepsilon=3$, \emph{AL-Guided-DP} decreases from $65.74\%$ to $35.95\%$ at the macro level, whereas the corresponding micro-level performance is considerably stronger. For EEG, \emph{DP-Post-AL} achieves higher macro defended ratios than \emph{AL-Guided-DP} at small $\varepsilon$, consistent with stronger privacy guarantees at low budgets, although this advantage is not uniform at larger $\varepsilon$. Similar patterns emerge for In-Location, in which CFs consistently erode micro-protection across $\varepsilon$.

\emph{Takeaway 4:} Beyond amplifying MIA accuracy (Takeaway~2) and weakening protection as $\varepsilon$ increases (Takeaway~3), CFs critically erode \emph{micro-level} protection, directly exposing true training members. While macro-level defended ratios may remain relatively high due to automatically saved records, the micro analysis reveals a substantial loss of protection for the most sensitive subset, the actual training data. These results indicate that aggregate exposure metrics can mask individual-level vulnerability, and that releasing CFs can significantly undermine the practical privacy guarantees of DP-based approaches.


\subsection{CF Quality Analysis}
Figure~\ref{Combined_CF} reports proximity and sparsity across scenarios. For EEG, proximity remains stable under DP (5.78–6.22 for \emph{Only-DP}; 5.60–7.15 for \emph{DP-Post-AL}), indicating that DP does not substantially degrade CF quality. Notably, \emph{AL-Guided-DP} exhibits slightly improved proximity compared to \emph{Only-AL}, which may be attributed to its use of a larger training subset. By incorporating more data points during training, \emph{AL-Guided-DP} may learn more stable decision boundaries, resulting in CFs that require smaller input changes. For In-Location, proximity values are generally higher than the \emph{Baseline}, reflecting the dataset's higher dimensionality. Sparsity follows a similar pattern: under tight privacy budgets (e.g., $\varepsilon=1$), DP slightly reduces sparsity, indicating stronger noise effects. In summary, proximity remains largely stable across privacy levels, whereas sparsity is mildly affected and exhibits greater sensitivity in high-dimensional settings.

Building on our analysis, \emph{Takeaway 5:} While DP introduces a utility–privacy trade-off (Takeaway~1) and CFs substantially amplify MIA success (Takeaways~2–4), the impact of DP on CF quality remains limited. Proximity remains stable across privacy budgets, and sparsity is only mildly affected, particularly in high-dimensional datasets.

These findings reveal a fundamental tension: explanations can significantly weaken privacy guarantees, yet privacy-preserving mechanisms can be applied without substantially degrading explanation quality. This creates a critical design challenge for MLaaS systems, which must carefully balance predictive utility, privacy robustness, and explanation fidelity. Beyond technical performance metrics, future work should examine the practical usefulness of CFs under privacy-preserving settings through empirical user studies. In particular, it is important to assess whether explanations that remain stable under DP also retain their interpretability and decision-support value for end users in high-stakes domains.

\section{Conclusion}\label{Conclusion}

In this paper, we investigate how counterfactual explanations (CFs) expand the attack surface of Machine Learning as a Service (MLaaS) systems by enabling highly effective shadow-based membership inference attacks (MIAs). We propose a defense framework that integrates Differential Privacy (DP) and Active Learning, and systematically analyze the resulting interplay between privacy protection, predictive performance, and explanation quality. Our results demonstrate that while DP can mitigate membership leakage, its protection can be substantially weakened when CFs are exposed. At the same time, we show that privacy-preserving training does not necessarily degrade explanation quality, revealing a fundamental tension between transparency and privacy resilience.

\bibliographystyle{IEEEtran}
{\footnotesize \bibliography{output}}

\end{document}